\documentclass[journal]{IEEEtran}
\usepackage{lineno,hyperref}
\usepackage{bbm}
\usepackage{bm}
\usepackage{amsfonts}
\usepackage{amsmath}
\usepackage{mathrsfs}
\usepackage{amssymb}
\usepackage{enumerate}
\usepackage{graphicx}
\usepackage{subfigure}
\usepackage{booktabs}
\usepackage{graphicx}
\usepackage{setspace}
\usepackage{tabularx}
\usepackage{algorithm}
\usepackage{algorithmic}
\usepackage{setspace}
\usepackage{multirow}
\usepackage{color}
\usepackage{cite}
\usepackage{array}
\usepackage{times}
\usepackage{mathptmx}
\usepackage{dsfont}
\ifCLASSINFOpdf
\else
\fi
\UseRawInputEncoding
\hypersetup{hypertex=true,
	colorlinks=true,
	linkcolor=blue,
	anchorcolor=blue,
	citecolor=blue}

\begin{document}

\title{Multi-Expert Adaptive Selection: Task-Balancing for All-in-One Image Restoration}

\author{Xiaoyan Yu, Shen Zhou,  Huafeng Li and Liehuang Zhu, \IEEEmembership{Senior Member, IEEE}
\thanks{This work was supported in part by the National Natural Science Foundation of China (No. 62161015), and the Yunnan Fundamental Research Projects (202301AV070004, 202401AS070106). (\textit{Corresponding
		author: Shen Zhou}.)}
\thanks{X. Yu is with the School of Computer Science and Technology, Beijing
		Institute of Technology, Beijing 100081, China (e-mail: xiaoyan.yu@bit.edu.cn).}
\thanks{S. Zhou and H. Li are with the Faculty of Information
	Engineering and Automation, Kunming University of Science and
	Technology, Kunming 650500, China.}
\thanks{L. Zhu is with the School of Cyberspace Science and Technology, Beijing
	Institute of Technology, Beijing 100081, China.}		
}

\markboth{Journal of \LaTeX\ Class Files,~Vol.~14, No.~8, August~2024}%
{Shell \MakeLowercase{\textit{et al.}}: A Sample Article Using IEEEtran.cls for IEEE Journals}


\maketitle

\begin{abstract}
The use of a single image restoration framework to achieve multi-task image restoration has garnered significant attention from researchers. However, several practical challenges remain, including meeting the specific and simultaneous demands of different tasks, balancing relationships between tasks, and effectively utilizing task correlations in model design. To address these challenges, this paper explores a multi-expert adaptive selection mechanism. We begin by designing a feature representation method that accounts for both the pixel channel level and the global level, encompassing low-frequency and high-frequency components of the image. Based on this method, we construct a multi-expert selection and ensemble scheme. This scheme adaptively selects the most suitable expert from the expert library according to the content of the input image and the prompts of the current task. It not only meets the individualized needs of different tasks but also achieves balance and optimization across tasks. By sharing experts, our design promotes interconnections between different tasks, thereby enhancing overall performance and resource utilization. Additionally, the multi-expert mechanism effectively eliminates irrelevant experts, reducing interference from them and further improving the effectiveness and accuracy of image restoration. Experimental results demonstrate that our proposed method is both effective and superior to existing approaches, highlighting its potential for practical applications in multi-task image restoration. The source code of the proposed method is available at \url{https://github.com/zhoushen1/MEASNet}.
\end{abstract}


\section{Introduction}
\IEEEPARstart{A}{s} an inverse problem in the field of computer vision, image restoration aims to restore high-quality and clear images from input images affected by various degradation factors such as haze, rain stains, noise, etc. Given its critical role in numerous downstream tasks, such as image fusion \cite{42, 45, 77}, target recognition \cite{43, 44}, and detection \cite{46,47}, this technology has attracted widespread attention from researchers. Although many methods have shown excellent performance in their respective fields, such as denoising \cite{2,8,84}, deblurring \cite{6,7,85}, deraining \cite{12,13,86,87}, and dehazing \cite{83,3,10,82}, they are often limited to dealing with a single type of degradation problem. When faced with different types of degradation or varying degrees of degradation, these methods often struggle to provide satisfactory results. To address the above challenges, researchers have begun exploring and designing universal models that can adapt to various image restoration tasks, achieving some promising research results \cite{21,23}. Although these methods perform well, their generality is still limited in terms of adaptability of the network model, and separate model training is still necessary for different restoration tasks. This means that such general frameworks are not suitable for handling multiple tasks simultaneously without sacrificing single-task performance.
\begin{figure}[t!]
	\centering
	\includegraphics[width=0.95\linewidth]{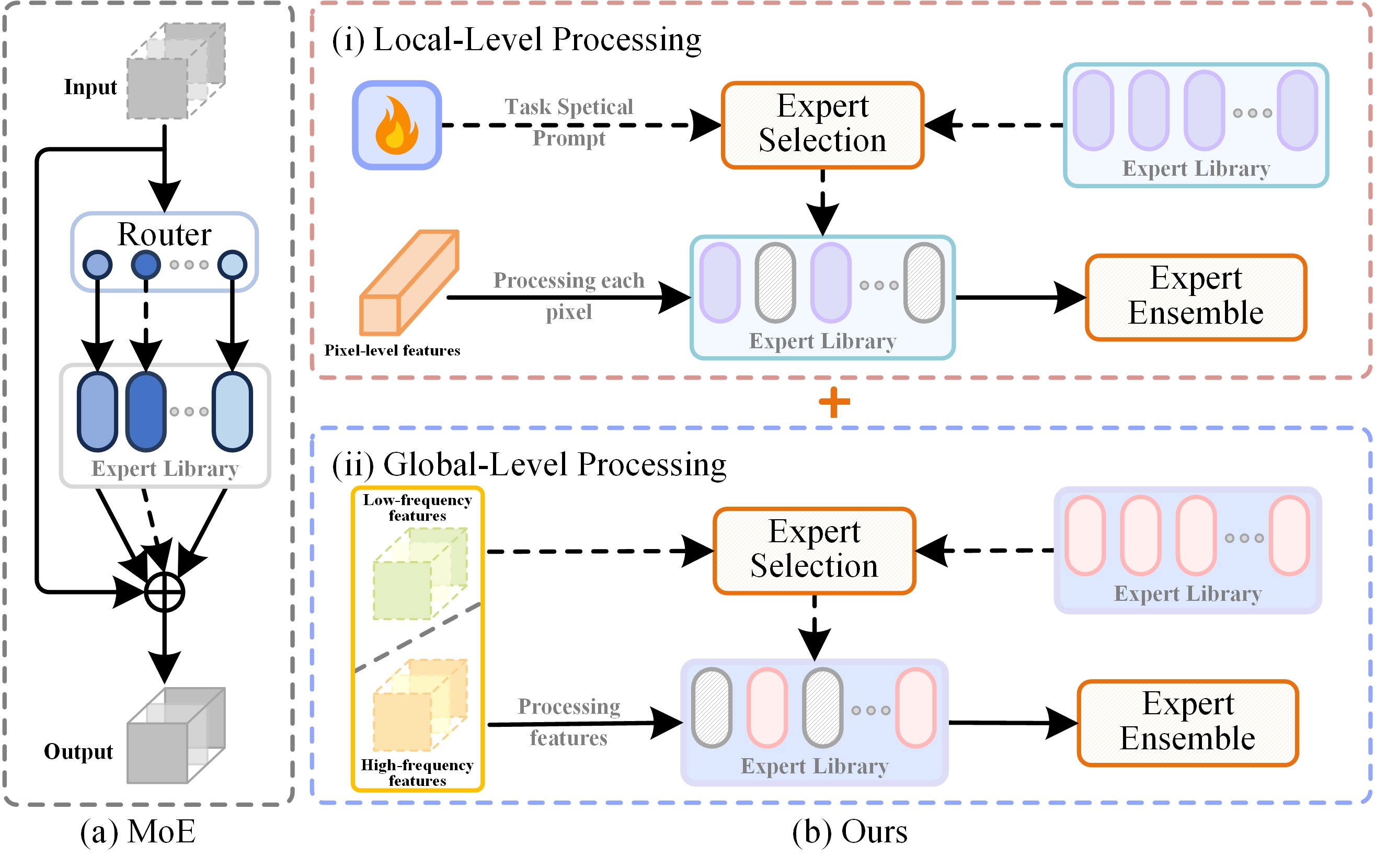}
	\vspace{-0.2cm}
	\caption{Comparison between the traditional Mixture of Experts (MoE) \cite{88} model and our proposed method. In MoE, the input is directly assigned to different experts for processing via a routing mechanism. The proposed method comprehensively considers the features at the pixel level as well as the global features composed of low-frequency and high-frequency components, and makes expert selection based on this.}\vspace{-0.3cm}
	\label{fig1}
\end{figure}

To solve the above problems, researchers have begun exploring the design of an All-in-One framework that can simultaneously handle multiple image restoration tasks with a single model. The key distinction from the aforementioned general model methods is that this framework can accommodate the requirements of multiple tasks without requiring retraining after the model is pre-trained. Therefore, the key to studying such methods lies in balancing the relationships between different tasks within the same framework. To solve these problems, researchers have begun exploring All-in-One frameworks designed to handle multiple image restoration tasks with a single model. Unlike general model methods, this framework can accommodate various tasks without requiring retraining after the model is pre-trained. Therefore, the key to designing such methods is to balance the relationships between different tasks within the same framework.  According to the characteristics of existing methods, current all-in-one image restoration methods can be roughly divided into three categories: Prior Knowledge-Based Image Restoration (PKIR) \cite{24,25,26,27,28,29}, Architecture Search and Feature Modulation-Based Image Restoration (AS-FM-IR) \cite{31,32,33,34,35,36}, and Prompt-Based Image Restoration (PromptIR) \cite{37,38,39,40,41}. PKIR  guides the image restoration process by utilizing external prior knowledge of images under different tasks, enabling the model to handle multiple types of image degradation simultaneously. However, such methods are limited by the accuracy and effectiveness of the prior knowledge of the input degraded image. If the prior knowledge does not match the characteristics of the actual image, the quality of the restored image may be significantly impacted.

Compared to PKIR, AS-FM-IR is also more prevalent in the ``All-in-One'' image restoration framework. In AS-FM-IR, architecture search-based methods aim to optimize relationships between different tasks by finding the most suitable network components for each task \cite{31,32,33}. Although these methods have shown effectiveness in practice, they require searching for network components based on the specific characteristics of the input image during deployment, which increases the complexity of model testing. On the other hand, feature modulation methods mainly generate a specific set of parameters based on the input image to adjust the network's output features to better adapt to each image restoration task \cite{34,35,36}. Although they have shown potential in addressing various types of image degradation problems, generating ideal modulation parameters remains a significant challenge when faced with complex and ever-changing real-world scenarios. The core idea of PromptIR methods is to dynamically generate a set of corresponding prompt information based on the input text prompts, using this information to assist the model in processing specific image restoration tasks more accurately \cite{37,38,39,40,41}. Through this approach, the model can achieve collaborative processing of multiple image restoration tasks with the assistance of prompts, ultimately improving overall processing efficiency and restoration quality. Although PromptIR techniques have shown potential and advantages, the effectiveness of such methods is still largely limited by the quality of generated prompts.

Although the aforementioned methods have made some research progress, achieving multi-task image restoration within a unified framework still faces numerous challenges. For instance, correlations among similar restoration tasks may exist, and leveraging these correlations effectively in model design to enhance overall performance and resource utilization efficiency remains a pressing concern. To tackle these challenges, we explore a multi-expert adaptive selection mechanism that coordinates different experts to cater to the diverse requirements of various tasks. To accomplish this objective, we devise a feature representation method based on the single pixel channel level of the image as well as the global level of the low-frequency and high-frequency components of the image. This method comprehensively captures the feature information of the image and provides robust support for subsequent selection and integration of experts.

Based on the aforementioned feature representations, we develop a multi-expert selection and ensemble scheme. As illustrated in Fig. \ref{fig1}, our proposed scheme exhibits notable distinctions from existing expert selection methods. It adaptively selects the most appropriate experts from the expert library for the current task and image content. This mechanism not only satisfies the individualized requirements of different tasks for network structures but also achieves a balance and optimization among tasks, ensuring that they can share expert resources without mutual interference. Furthermore, since expert selection is contingent upon image content, different tasks can establish correlations via shared experts. This mechanism enables the model to share knowledge and learn from each other when tackling diverse tasks, ultimately enhancing the overall restoration effect. In this process, the multi-expert mechanism also has the capability to eliminate irrelevant expert models, effectively mitigating interference between tasks. Experimental results demonstrate that the method proposed in this paper is not only effective but also possesses significant advantages over existing methods, exhibiting superior restoration performance and enhanced generalization ability in the domain of multi-task image restoration.

In summary, the contributions and advantages of the proposed method are primarily reflected in the following aspects:

\begin{itemize}
	\item A multi-expert adaptive selection mechanism is designed to adaptively select the most suitable experts for the current task and image content, taking into account the characteristics of the input image and the prompts of specific tasks. This design not only meets the personalized requirements of different tasks for network structures but also achieves balance and optimization among tasks, ensuring that they can coordinate and share network resources without interfering with each other.
	
	\item Considering the complementarity between local and global image features, we design a feature representation method that integrates the channel-level information of individual pixels with the global-level low-frequency and high-frequency components. This innovative approach captures image features more comprehensively, providing strong support for subsequent expert selection and ensemble processes.
	
	\item Since expert selection is contingent upon image content, associations between different tasks are established through the sharing of experts. This mechanism enables the model to learn from and share knowledge across diverse tasks, ultimately enhancing overall performance and improving the efficiency of expert utilization.
	
	\item This multi-expert mechanism also possesses the ability to exclude experts that are irrelevant to the current task, effectively mitigating interference between tasks and further enhancing the effectiveness and accuracy of image restoration. Experimental results demonstrate that, compared to existing methods, this approach exhibits superior restoration performance and improved generalization ability in the field of multi-task image restoration.
\end{itemize}

The remaining content of this paper is arranged as follows: Section \ref{R2}
 reviews the methods related to the paper. Section \ref{R3} provides a detailed introduction to the proposed method. Section \ref{R4} presents the experimental results and analysis. Finally, Section \ref{R5} summarizes and discusses the proposed method.

\section{Related Work}\label{R2}
\subsection{Task-Specific Image Restoration}
Image restoration is a significant topic in the field of computer vision. Due to the complexity and uncertainty of the degradation process, traditional image restoration methods often rely on manually designed features and prior knowledge to construct restoration models \cite{1,2,3,4,5}. While these methods can achieve pleasing results on specific datasets, their performance is often limited when dealing with more diverse and complex degraded images in the real world. With the rapid development of deep learning, image restoration based on convolutional neural networks (CNNs) has received widespread attention \cite{16,6,8,10,17,11}. Numerous effective methods have been proposed for various tasks, including image deblurring \cite{6,7}, image denoising \cite{8,9}, image dehazing \cite{10,11,18}, image deraining \cite{12,13}, and image desnowing \cite{14,15}. Representative methods include DnCNN for denoising \cite{8}, MSPF for deraining \cite{12}, DCMPNet for dehazing \cite{10}, and DTCW for desnowing \cite{15}. However, CNNs have limitations in modeling long-range dependencies in images. To address this issue, researchers have introduced Transformers \cite{78} to the field of image restoration, proposing methods such as SwinIR \cite{22} and Restormer \cite{23}. These Transformer-based methods can better restore details and clarity in images with complex textures and structures.

However, existing deep learning-based methods, while performing well on specific tasks, still show certain limitations when dealing with different types of image degradation using a unified framework. Each type of image degradation has its unique characteristics, requiring targeted network designs and optimization strategies. To enhance the generality of these methods, some studies have begun to explore new model design strategies. For instance, Wang et al. proposed Uformer \cite{21}, Liang et al. proposed SwinIR \cite{22}, and Zamir et al. proposed Restormer \cite{23}. These methods further enhance the performance and stability of image restoration by introducing new network structures and optimization algorithms. Although these methods have shown effectiveness in various image restoration tasks, addressing the specific needs of each task within a unified framework remains challenging. For example, denoising tasks may focus more on restoring local textures, while deblurring tasks may emphasize restoring global structures. This indicates that achieving multi-task recovery within a single framework, while preventing the performance of individual task recovery from degrading, remains a significant challenge. 
\begin{figure*}[t!]
	\centering
	\includegraphics[width=0.92\linewidth]{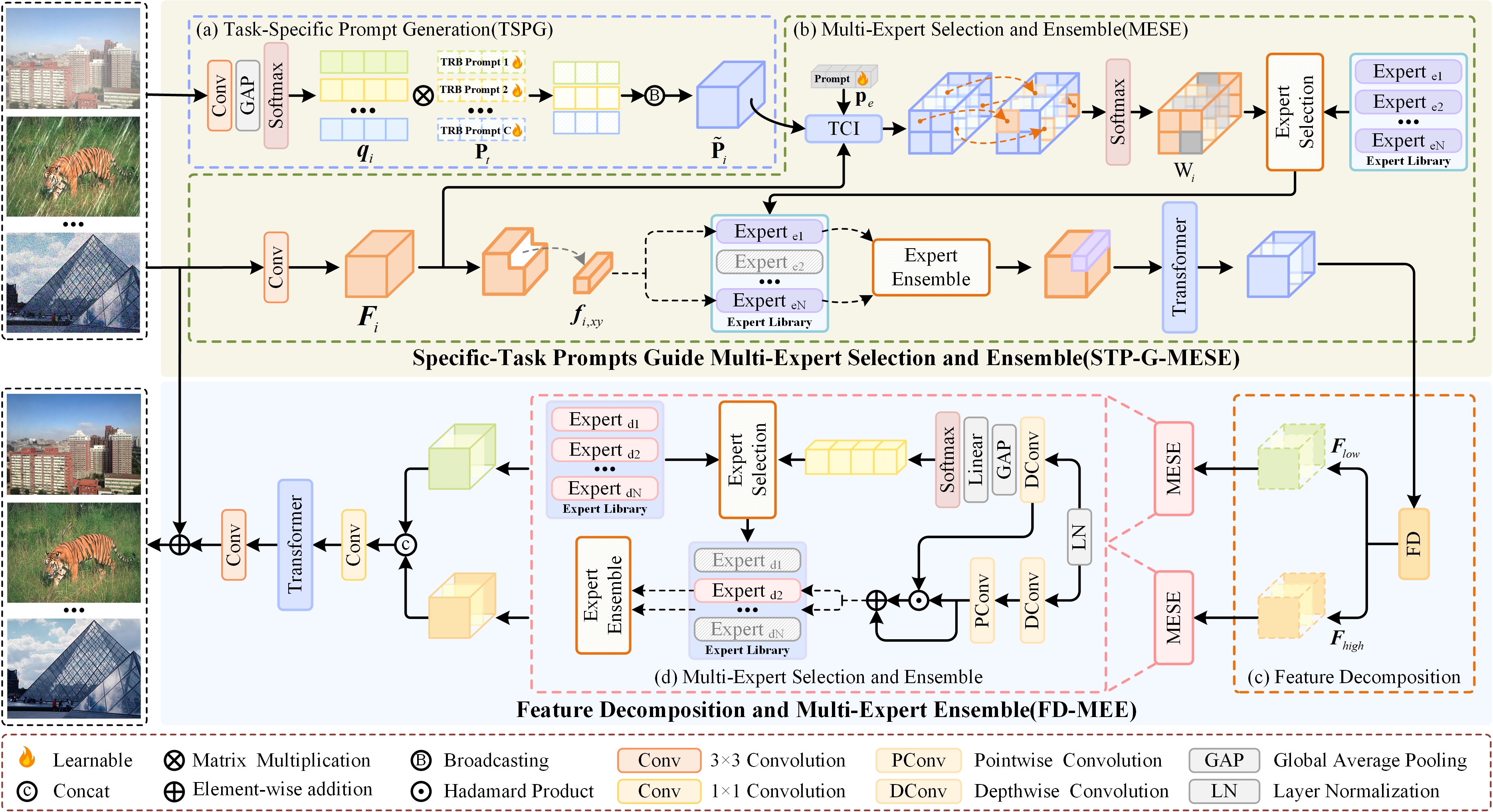}
	\vspace{-0.2cm}
	\caption{Overview of the proposed method: The input degraded image is first processed by TSPG, and task-related prompts are generated with the assistance of TRB Prompt. Subsequently, effective integration of task and content-related information is achieved through TCI. Based on the information integrated by TCI, we predict the importance of each expert in the expert database for the current sample pixel restoration task and select the top-$K$ most important experts to handle each pixel in the image. In the stage of feature decomposition and multi-expert ensemble, the FD separates the features output by the Transformer layer into high-frequency and low-frequency components, allowing specific experts to restore the image information as a whole for different frequency components. }\vspace{-0.3cm}
	\label{fig2}
\end{figure*}

\subsection{All-in-One Image Restoration}
\subsubsection{Image Restoration via Knowledge-Based Methods}
In All-in-One image restoration frameworks, leveraging external knowledge is a common strategy for addressing conflicts arising from the inconsistent demands of diverse restoration tasks on a unified framework \cite{24,25,26,27,28,29}. Specifically, Chen et al. \cite{24} transferred the task-specific prior knowledge contained in multiple single-task teacher networks to a student network via knowledge transfer and contrastive regularization, enabling compatibility with various degraded image restoration tasks. Wang et al. \cite{25}, on the other hand, utilized a codebook trained on high-quality images to replace the features of degraded regions in images, effectively addressing the incompatibility among specific requirements for the restoration model. Zhang et al. \cite{30} revealed the intrinsic connections between multiple types of degradation by employing the underlying physical principles of different degradation types and a learnable principal component analysis method. They further utilized these connections to construct a dynamic routing mechanism to remove unknown degradation. Lastly, Jiang et al. \cite{26} constructed a correlation between degraded images and predefined degradation descriptions, retrieved the text descriptions of degraded images, and used them as generation conditions for diffusion models, ultimately achieving multi-task image restoration.

Similarly, Lin et al. \cite{27} proposed a method that maps degraded images into a text space. They obtained text descriptions of clear images by removing degradation-related words and used these descriptions as generation conditions for diffusion models. This approach achieved joint restoration of images with various degradation types. Tan et al. \cite{28}, on the other hand, incorporated the CLIP Weather Prior embedding module into their image restoration model. This allowed the model to extract prior information related to specific degradation types from input samples using the CLIP image encoder. Based on this information, the model could dynamically adjust its internal parameters, effectively addressing and restoring various types of image degradation. Luo et al. \cite{29} took a different approach by fine-tuning the CLIP image encoder to predict high-quality feature embeddings from various degraded images. They used this embedding as a generation constraint for diffusion models, achieving the joint execution of different degradation restoration tasks. However, despite their effectiveness, these methods rely heavily on external prior knowledge. This reliance may reduce their flexibility in facing unknown or changing conditions to some extent.

\subsubsection{Image Restoration via ArchSearch and Feature Modulation}
In addition to the above methods, architecture search (ArchSearch) and feature modulation are also quite common in the 'all-in-one' image restoration framework. ArchSearch effectively addresses the issue of inconsistency between different task requirements in the overall network framework design. Specifically, Chen et al. \cite{31} constructed a set of decoders tailored for specific tasks. They intelligently searched for and selected the most appropriate decoder based on the degradation type of the input image to reconstruct the restored image, significantly enhancing the model's ability to handle a wide range of degraded images. Park et al. \cite{32} proposed a multi-degradation adaptive classifier and utilized it to select suitable filters as convolution kernels for efficient feature extraction. Zhu et al. \cite{33} employed a two-stage training strategy. This strategy initially learns the general features of degradation and then delves into specific features. Based on this, it searches for the convolutional layer that best matches the current task from a pool of convolutional layers, enabling the model to flexibly adapt to various types of degraded images.
Yang et al. \cite{79} constructed an expert library containing various convolution kernels. They introduced a question-answering model to extract degradation information and locate degradation positions from input images. This information then guides the selection process of convolution kernels. Although these methods are effective, the search process is relatively complex, which hinders the practical deployment and application of the model.

In terms of feature modulation, Li et al. \cite{50} employed a contrastive learning strategy to learn degradation representations from degraded images. These representations were then used to modulate features within the restoration network. Wei et al. \cite{34} effectively extracted information on degradation types and severity by combining edge quality ranking loss with contrast loss. This approach enabled the generation of parameters for affine transformation of features, achieving affine modulation. Cui et al. \cite{35} utilized a frequency mining module for spectral decomposition of degraded images, extracting both low-frequency and high-frequency components. These components were then modulated through a frequency modulation module. Chen et al. \cite{36} pre-trained a general image restoration model using synthesized degraded images. The model was then fine-tuned with adapters to meet the requirements of specific image restoration tasks. Although these methods have demonstrated effectiveness in addressing various types of degradation, they often struggle to generate ideal modulation parameters in more complex situations. This limitation can lead to modulation outcomes that fail to meet specific task requirements.

\subsubsection{Image Restoration via Prompt Learning}
With advancements in prompt learning research, researchers have begun exploring its use to address the challenge of coordinating multiple tasks within a unified framework. Specifically, Potlapalli et al. \cite{37} introduced learnable prompts to implicitly capture various types of degradation information and modulate features accordingly. Ma et al. \cite{38} employed degradation-aware visual prompts to encode different types of image degradation information, using linear weighting to control the restoration process. Li et al. \cite{39} integrated degradation-aware prompts and restoration prompts into a general restoration prompt and utilized a prompt-feature interaction module to modulate degradation-related features. Kong et al. \cite{40} adopted a sequential learning strategy to optimize restoration objectives and resolve conflicts during training, enhancing the network's adaptability to different restoration tasks through an explicit adaptive prompting mechanism. Marcos et al. \cite{41} proposed a text-guided image restoration model that uses human instructions as prompts to guide the restoration process. While prompt-based image restoration methods have demonstrated some effectiveness, they are often limited by the quality of the generated prompts. Unlike these methods, our approach leverages adaptive selection and comprehensive utilization of multiple experts. It considers both the connections and differences between various tasks, ultimately achieving superior performance in joint restoration.

\section{The Proposed Method}\label{R3}
\subsection{Overview}
Our core goal is to build an All-in-One image restoration framework that can effectively recover clear images from degraded images without relying on prior information about the degradation of the input image. As shown in Fig. \ref{fig2}, the proposed framework consists of two main components: (i) Task-Specific Prompt-Guided Multi-Expert Selection and Ensemble (STP-G-MESE) and (ii) Feature Decomposition and Multi-Expert Ensemble (FD-MEE). The STP-G-MESE component primarily selects the most suitable expert for the restoration of the current pixel based on features across different channels. This module adaptively generates task-related prompts based on the degradation of the input image and selects the most appropriate experts for the current image based on these prompts and image content, effectively mitigating interference from different tasks.

In contrast, FD-MEE extracts image features from a global perspective. It fully explores the information contained within the entire image to achieve a comprehensive representation of the global features. Technically, FD-MEE decouples the input features into high-frequency and low-frequency components and selects the most suitable experts from the expert library based on these components. This approach allows the features of different frequency components to play their respective roles in the restoration process.
Our method integrates pixel-level features across different channels and global image features within a single framework, resulting in an effective representation of image content. Notably, the experts used in this framework consist of multi-layer perceptrons (MLPs) with varying parameters, which endows the framework with robust learning and adaptation capabilities.

\subsection{Specific-Task Prompts Guide Multi-Expert Selection and Ensemble}
STP-G-MESE mainly consists of two core components: Task-Specific Prompt Generation (TSPG) and Expert Selection and Ensemble. The TSPG component generates prompt information that is highly relevant to the task from the input image, providing a key basis for subsequent expert selection. Guided by TSPG, the Expert Selection and Ensemble component identifies the most suitable experts from the expert library for the current sample recovery task. The features output by these selected experts are then integrated to comprehensively leverage their strengths and enhance image quality restoration.

\subsubsection{Task-Specific Prompt Generation }
To effectively handle various types of image restoration tasks within a unified framework, this paper proposes an innovative strategy that utilizes multi-expert collaboration. This strategy aims to coordinate and address the unique requirements of each task within the model. Selecting an appropriate expert network to assist with specific image restoration tasks is crucial for the overall performance of the model. Indeed, the performance of a trained model is influenced not only by the type of task but also by the specific content of the image. To comprehensively address the impact of these two core factors on image restoration performance, this paper designs a TSPG mechanism. This mechanism generates prompt information that is closely related to specific tasks. The detailed implementation process is illustrated in Fig.\ref{fig2}(a).

\begin{figure}[!t]
	\centering
	\includegraphics[width=0.8\linewidth]{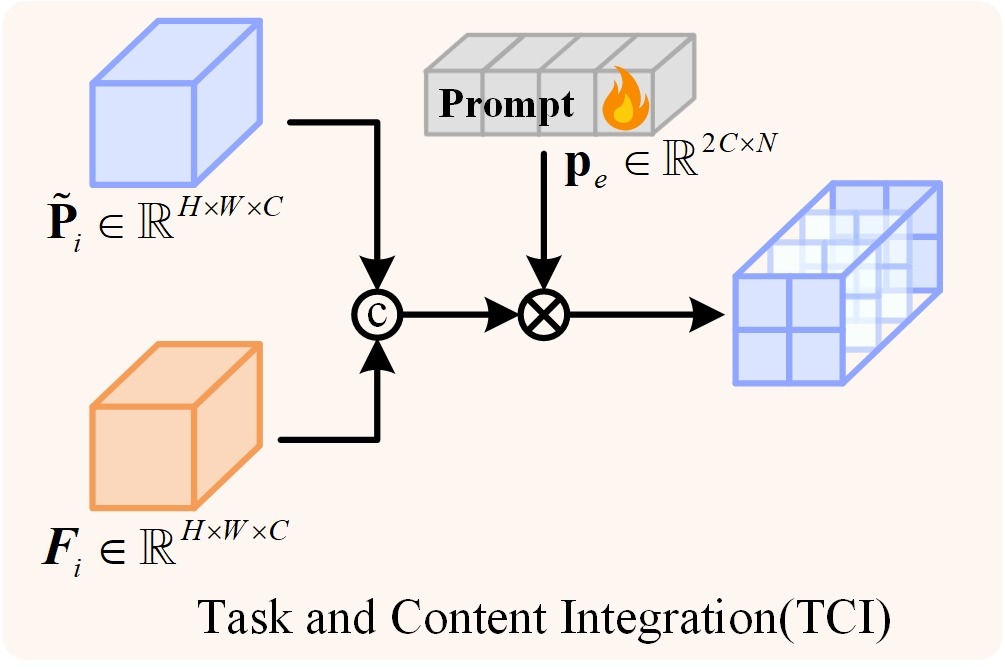}\vspace{-0.2cm}
	\caption{Framework structure diagram of TCI.}\vspace{-0.3cm}
	\label{fig3}
\end{figure}
Let the image to be restored be $\mathbf{\tilde{u}}_i^{*}$. In the Task-Specific Prompt Generation (TSPG) module, $\mathbf{\tilde{u}}_i^{*}$ first undergoes convolution processing, followed by global average pooling, and finally passes through the Softmax layer to produce $\mathbf{q}_i \in \mathbb{R}^{1 \times C}$. Here, $C$ represents the number of channels in the feature map output of the convolutional layers. To address the specific needs of image restoration, we introduce a set of task-specific and learnable prompts to assist in expert selection. While prompts can be explicitly defined for each task, different image restoration tasks are interrelated rather than independent. Therefore, this interrelation must be thoroughly considered when designing the task-specific prompt generation mechanism.

To address this issue, we propose a method for generating prompts that meet specific task requirements from a set of Task-Related Basic Prompts (TRB Prompts). Specifically, we use the output of the Softmax layer, $\mathbf{q}_i$, as the weight for the basic prompts related to the task, to combine and construct prompts for this specific task. Assuming the basic prompts related to the task are $\mathbf{P}_t = [\mathbf{p}_{t1}, \mathbf{p}_{t2}, \cdots, \mathbf{p}_{tC}]^T \in \mathbb{R}^{C \times M}$, the prompts for the input image $\tilde{u}_i^*$ can be expressed as:
\begin{equation}
	\begin{aligned}
		\mathbf{\tilde{p}}_i = \mathbf{q}_i \times \mathbf{P}_t
	\end{aligned}
\end{equation}
where $M$ is the length of $\mathbf{p}_{tc}$ ($c = 1, 2, \cdots, C$). For ease of calculation in the subsequent process, we set $M$ to be equal to the number of channels $C$ of the feature maps output by the convolutional layer. We then broadcast $\mathbf{P}_t$ as $\mathbf{\tilde{P}}_i \in \mathbb{R}^{H \times W \times C}$ to match the feature map output by the convolutional layer in the multi-expert selection and ensemble module.
\subsubsection{Multi-Expert Selection and Ensemble}
The performance of the model is not only affected by the task category but also deeply influenced by the characteristics of the image content itself. Therefore, when selecting suitable experts for different tasks, we need to carefully consider both the characteristics of the tasks and the uniqueness of the image content. Since $\mathbf{q}_i$ is the result obtained by applying global average pooling to the Softmax layer, it contains information about image degradation. Thus, the $\mathbf{\tilde{P}}_i$ created based on $\mathbf{q}_i$ naturally embeds this degradation information, allowing $\mathbf{\tilde{P}}_i$ to serve as prompts for specific tasks in the image restoration process. Additionally, the image features output by the convolutional layer in the Multi-Expert Selection and Ensemble (MESE) module are rich in specific content information of the image and can serve as prompts for image content, aiding in the more accurate selection of suitable experts.

Specifically, we concatenate $\mathbf{\tilde{P}}_i$ and $\mathbf{F}_i$ to obtain $\mathbf{F}_{c, i} = [\mathbf{\tilde{P}}_i, \mathbf{F}_i] \in \mathbb{R}^{H \times W \times 2C}$, where $[\cdot]$ represents the concatenation operation. To select suitable experts to handle pixels at different positions, we need to synthesize $\mathbf{F}_{c, i}$ along the channel dimension to evaluate which expert should be selected for quality restoration of pixels at each position. To achieve this goal, we introduce learnable expert prompts $\mathbf{p}_e \in \mathbb{R}^{2C \times N}$, where $N$ represents the total number of experts required. Using these expert prompts, the features are integrated along the channel, and the integrated results are fed into the Softmax function to obtain the demand for each expert for pixels at different positions. As shown in Fig. \ref{fig3}, this process can be formulated as:
\begin{equation}
	\begin{aligned}
		\mathbf{W}_i = \text{Softmax}(\mathbf{F}_{c,i} \times \mathbf{p}_e)
	\end{aligned}
\end{equation}

The pixel value at position $(x,y)$ in the $n$-th channel of $\mathbf{W}_i$ represents the degree to which the pixel at that position requires the $n$-th expert. The higher the value, the greater the demand for the $n$-th expert to restore that pixel. Instead of using all experts based on their demand degrees, we select the top $K$ experts with the highest demand degrees to participate in pixel feature recovery. This approach effectively enhances the feature recovery ability of the selected experts and reduces interference between unrelated tasks. Additionally, because the same experts may be selected for correlated but different image restoration tasks, these tasks can establish associations through shared experts. This association is beneficial for improving the performance of the current task by leveraging insights from other related tasks.

Let $\mathbf{f}_{i,xy} \in \mathbb{R}^{1 \times C}$ denote the feature of the pixel at position $(x, y)$ in $\mathbf{F}_i$, and let the expert applicable to $\mathbf{f}_{i,xy}$ be denoted as $\mathbf{E}_{i,xy}^{j_k}$, where ${j_k} \in {1, \cdots, N}$ and $k = 1, \cdots, K$. The processed features are represented as $\mathbf{f}_{i,xy}^{j_k} = \mathbf{E}_{i,xy}^{j_k}(\mathbf{f}_{i,xy})$. To reflect the importance of different experts, we apply the weights ${\mathbf{w}}_{i,xy}^{j_k}$ stored in ${\mathbf{W}}_i$ for the expert $\mathbf{E}_{i,xy}^{j_k}$ to modulate the feature $\mathbf{f}_{i,xy}^{j_k}$:
\begin{equation}
	\begin{aligned}
		\mathbf{f}_{i,xy} = \sum_{k=1}^K {\mathbf{w}}_{i,xy}^{j_k} \mathbf{f}_{i,xy}^{j_k}
	\end{aligned}.
\end{equation}
Since the above operation only considers the spatial correlation of the pixel and does not account for the correlation between pixel feature vectors, we add a transformer layer after the multiple experts to explore the correlation between different pixels.
\begin{figure}[!t]
	\centering
	\includegraphics[width=0.85\linewidth]{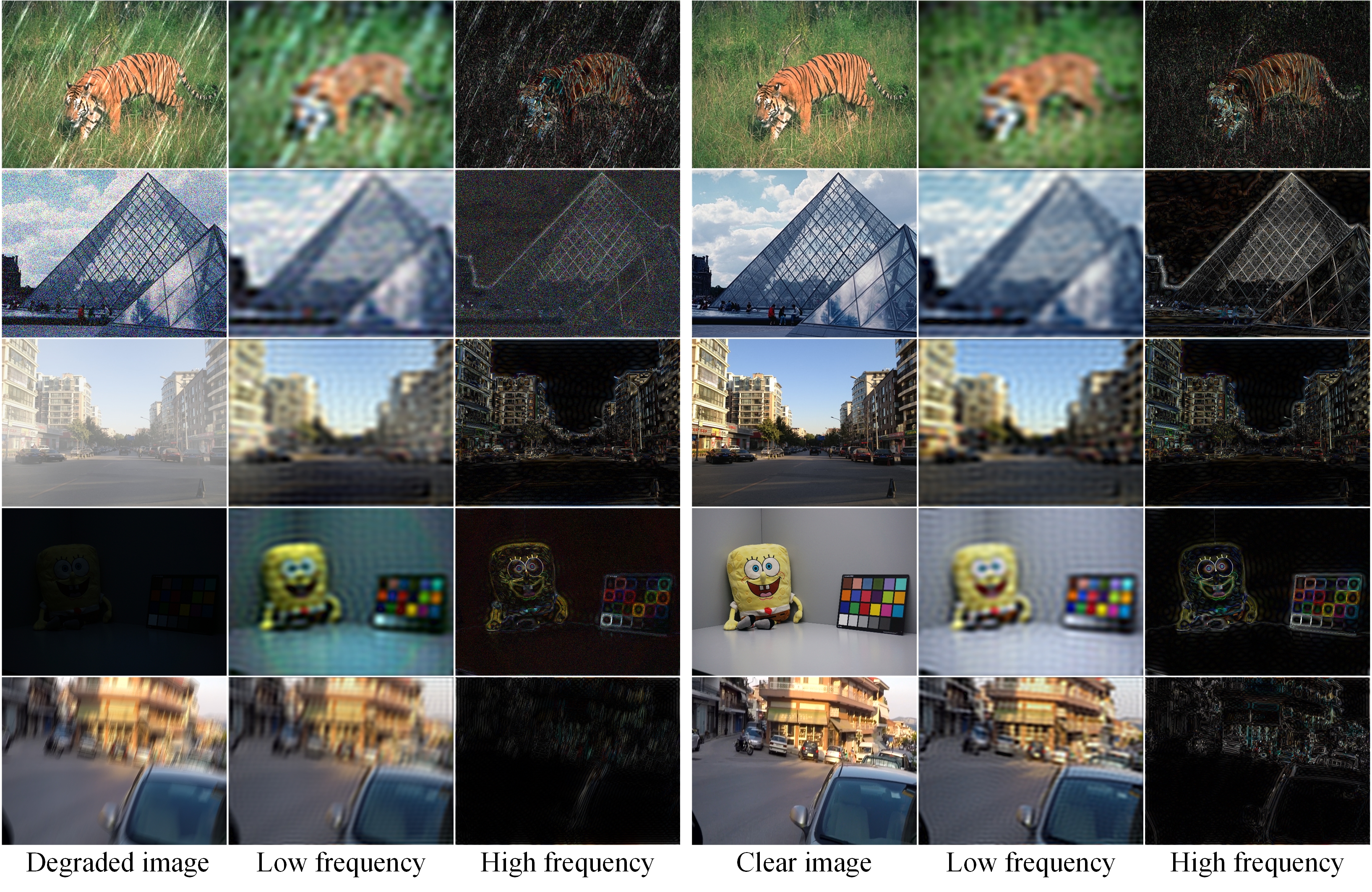}\vspace{-0.2cm}
	\caption{Influence of different degradation types on image high- and low-frequency components.}\vspace{-0.3cm}
	\label{fig4}
\end{figure}

According to the above process, the importance level of the $n$-th expert to all pixels on a feature map of size $H \times W$ can be described by summing all the values on the $n$-th channel of $\mathbf{W}_i$:
\begin{equation}
	\begin{aligned}
			S_i(n) = \sum\limits_{x = 1}^W \sum\limits_{y = 1}^H \mathbf{W}_i(n, x, y), \quad n = 1, 2, \cdots , N 
		\end{aligned}.
\end{equation}		
Assuming that the pixel at $(x, y)$ selects the $n$-th expert, we set the value at the $n$-th channel position $(x, y)$ in $\mathbf{W}_i$ to 1, and treat other positions where the $n$-th expert is not selected as 0, thus forming a new tensor $\mathbf{\tilde{W}}_i \in \mathbb{R}^{W \times H \times N}$. The total number of times the $n$-th expert is selected can be expressed as:
\begin{equation}
	\begin{aligned}
			\tilde{S}_i(n) = \sum\limits_{x = 1}^W \sum\limits_{y = 1}^H \mathbf{\tilde{W}}_i(n, x, y)
	\end{aligned}.
\end{equation} 
Let the mean and standard deviation of the sequences $\{S_i(n)\}_{n=1}^N$ and $\{\tilde{S}_i(n)\}_{n=1}^N$ be $\left(\mu_{S_i}, \sigma_{S_i}\right)$ and $(\mu_{\tilde{S}_{i}}, \sigma_{\tilde{S}_{i}})$, respectively. To ensure that all experts can be selected by pixels with the same probability, we introduce the coefficient of variation squared loss \cite{80} to optimize the relevant parameters:
\begin{equation}
	\begin{aligned}
			\ell_{balance} = \frac{\sigma_{S_i}}{\mu_{S_i}^2 + \varepsilon} + \frac{\sigma_{\tilde{S}_i}}{\mu_{\tilde{S}_i}^2 + \varepsilon}
	\end{aligned},
\end{equation} 
where $\varepsilon$ is an extremely small positive number used to avoid the denominator being zero.\vspace{-0.2cm}

\subsection{Feature Decomposition and Multi-Expert Ensemble}
\begin{figure}[!t]
	\centering
	\includegraphics[width=0.8\linewidth]{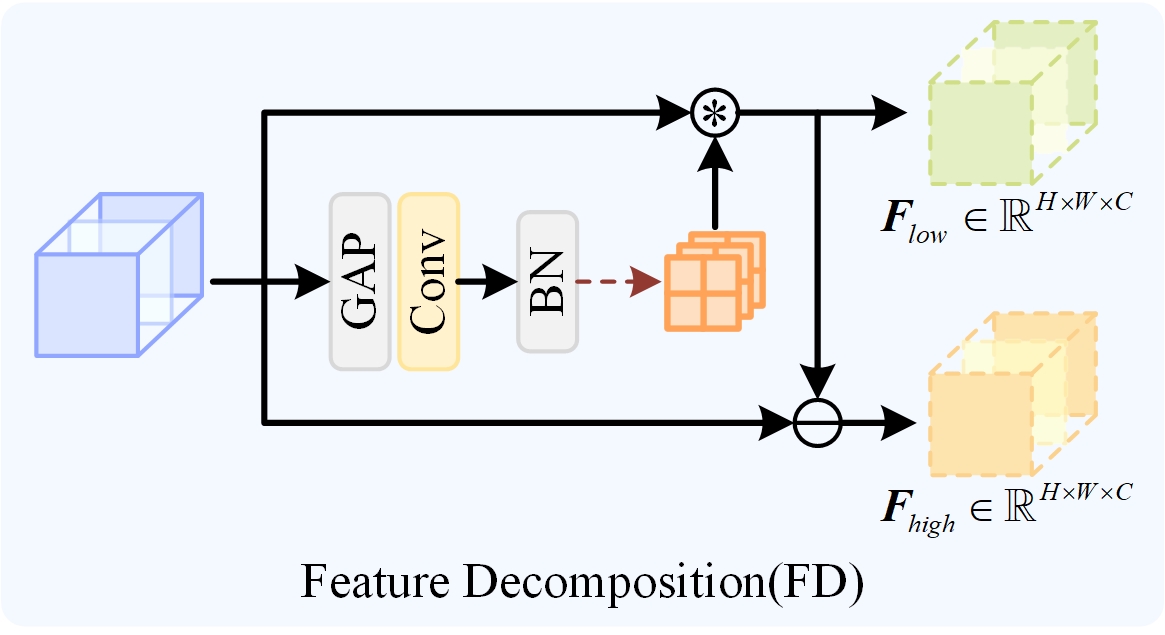}
	\vspace{-0.3cm}
	\caption{Framework structure diagram of FD.}\vspace{-0.3cm}
	\label{fig5}\vspace{-0.3cm}
\end{figure}

In reality, each type of image degradation affects the content of the image in a specific way. As shown in Fig.\ref{fig4}, noise and haze mainly affect the high-frequency components of the image, low light conditions primarily affect the low-frequency components, and rain marks and blurring affect both the high-frequency and low-frequency components. Therefore, when restoring the quality of degraded images, processing the low-frequency and high-frequency components separately can reduce mutual interference during the restoration process and make the processing more targeted. To achieve effective separation of high-frequency and low-frequency components, this paper introduces dynamically learnable filters \cite{81}. These filters can dynamically learn information from each spatial position and channel, demonstrating excellent feature separation capabilities. Assuming that ${\mathbf{F}_{i,t}}$ is the output of the Transformer layer in STP-G-MSE, in the FD of FD-MEE (as shown in Fig.\ref{fig5}), the result ${\mathbf{G}_{i,low}}$ obtained by sequentially passing through global average pooling, convolutional layers, batch normalization, and Softmax is used as a low-pass filter:
\begin{equation}
	\begin{aligned}
{\mathbf{G}_{i,low}} = \rm{Softmax}\left( {{\rm{BN}}\left( {\rm{Con}{v_{1 \times 1}}\left( {\rm{GAP}\left( {{{\bf{F}}_{i,t}}} \right)} \right)} \right)} \right)
	\end{aligned},
\end{equation} 
where $\mathbf{G}_{i,low}$ and $\rm{Conv}_{1 \times 1}$ represent the global average pooling layer and the 1$\times$1 convolution operation, respectively; BN represents batch normalization processing. Softmax is used to ensure that the generated filter is a low-pass filter. After obtaining $\mathbf{G}_{i,low}$, we use the operations in formulas (7) and (8) to separate high-frequency and low-frequency features.
\begin{equation}
	\begin{aligned}
    &\mathbf{F}_{i,low} = \mathbf{F}_{i,low}*{\mathbf{F}_{i,t}},\\&
    \mathbf{F}_{i,high} = \mathbf{F}_{i,t} - \mathbf{F}_{i,low},
	\end{aligned}
\end{equation} 
where $*$ represents convolution operation.

After obtaining $\mathbf{F}_{i, low}$ and $\mathbf{F}_{i, high}$, we feed them into the multi-expert ensemble module to select experts from the global level of low-frequency and high-frequency components of the input image. This compensates for the shortcomings of only achieving image restoration from a single pixel as discussed in section 3.2. As shown in Fig.\ref{fig2}, the multi-expert selection and ensemble at the global level consists of two branches: upper and lower. The upper branch is used to generate the identity information of experts, while the lower branch is used to achieve the multi-expert ensemble.
In the upper branch, $\mathbf{F}_{i, low}$ ($\mathbf{F}_{i, high}$) sequentially passes through LN, DConv, GAP, and Linear layers. The obtained result is processed by Softmax to predict the expert information and store it in the degree vector $\mathbf{w}_{i, low} \in \mathbb{R}^{1 \times N}$ ($\mathbf{w}_{i, high} \in \mathbb{R}^{1 \times N}$). The $n$-th dimensional data in $\mathbf{w}_{i, low}$ ($\mathbf{w}_{i, high}$) represents the importance of the $n$-th expert $\mathbf{E}_{i, low}$ ($\mathbf{E}_{i, high}$) to the image $\mathbf{\tilde{u}}_i^*$. We select the top $K$ most important experts $\mathbf{E}_{i, low}^{j_k}$ ($\mathbf{E}_{i, high}^{j_k}$) for $\mathbf{\tilde{u}}_i^*$ based on $\mathbf{w}_{i, low}$ ($\mathbf{w}_{i, high}$), where $j_k \in \{1, \cdots, N\}$ and $k \in \{1, \cdots, K\}$.

As DConv extracts features from a single feature map, while PConv extracts features from the channel dimension, DConv captures the spatial relationships between pixels, whereas PConv captures the correlations between features in different channels. This creates a distinction between the features extracted by DConv and PConv. Therefore, the interaction between the outputs of DConv and PConv can be utilized to highlight the role of key information in image restoration. Assuming that in the lower branch, $\mathbf{F}_{i, low}$ ($\mathbf{F}_{i, high}$) passes through LN, DConv, and PConv in sequence, the result obtained is $\mathbf{\tilde{F}}_{i, l}^p$ ($l = \text{low, high}$). In the upper branch, the output result of DConv is $\mathbf{\tilde{F}}_{i, l}^d$ ($l = \text{low, high}$). We use formula (9) to implement the interaction between $\mathbf{\tilde{F}}_{i, l}^p$ and $\mathbf{\tilde{F}}_{i, l}^d$ ($l = \text{low, high}$):
\begin{equation}
	\begin{aligned}
		\mathbf{\tilde F}_{i,l}^{pd} = \mathbf{\tilde F}_{i,l}^p \odot \mathbf{\tilde F}_{i,l}^d + \mathbf{\tilde F}_{i,l}^p ~~(l = low,high)		
	\end{aligned}
\end{equation} 

We feed $\mathbf{\tilde{F}}_{i,l}^{pd}$ into the selected $K$ experts $\mathbf{E}_{i,l}^{j_k}$, where $k \in \{1, \cdots, K\}$, and the result is expressed as $\mathbf{F}_{i,l}^{j_k} = \mathbf{E}_{i,l}^{j_k}(\mathbf{\tilde{F}}_{i,l}^{pd})$. To reflect the different roles played by various experts in image restoration tasks, we adopt the following approach to achieve effective ensemble of multiple experts:
\begin{equation}
	\begin{aligned}
		\mathbf{\tilde{F}}_{i,l}= \sum_{k = 1}^K \mathbf{w}_{i,l}(j_k) \times \mathbf{\tilde{F}}_{i,l}^{j_k}
	\end{aligned}
\end{equation}
where $\mathbf{w}_{i,l}(j_k)$ is the $j_k$-th element in the vector $\mathbf{w}_{i,l}$, which represents the importance of the $j_k$-th expert to $\mathbf{\tilde{u}}_i^*$. 
 
To achieve effective restoration of image information, we concatenate $\mathbf{F}_{i,low}$ and $\mathbf{F}_{i,high}$ and feed them into a decoder consisting of convolutional layers, transformer layers, and convolutional layers cascaded to reconstruct the restoration result. We then add the restoration result to the original image to reconstruct the final restored image $\mathbf{\tilde{u}}_i$. To ensure the quality of the recovery results, we use $l_1$-loss to optimize network parameters:
\begin{equation}
	\begin{aligned}
     {\ell _1} = {\left\| {\mathbf{\tilde u}}_i - \mathbf{u}_i \right\|_1}	
	\end{aligned}
\end{equation}
where $\mathbf{u}_i$ is the ground truth corresponding to ${\mathbf{\tilde u}}_i$. The total loss used for model training is:
\begin{equation}
	\begin{aligned}
    {\ell _{total}} = {\ell _1} + \lambda {\ell _{balance}}
	\end{aligned}
\end{equation}
where $\lambda$ is a hyperparameter used to adjust the role played by $\ell_{balance}$.


\begin{figure*}[t!]
	\centering
	\includegraphics[width=0.92\linewidth]{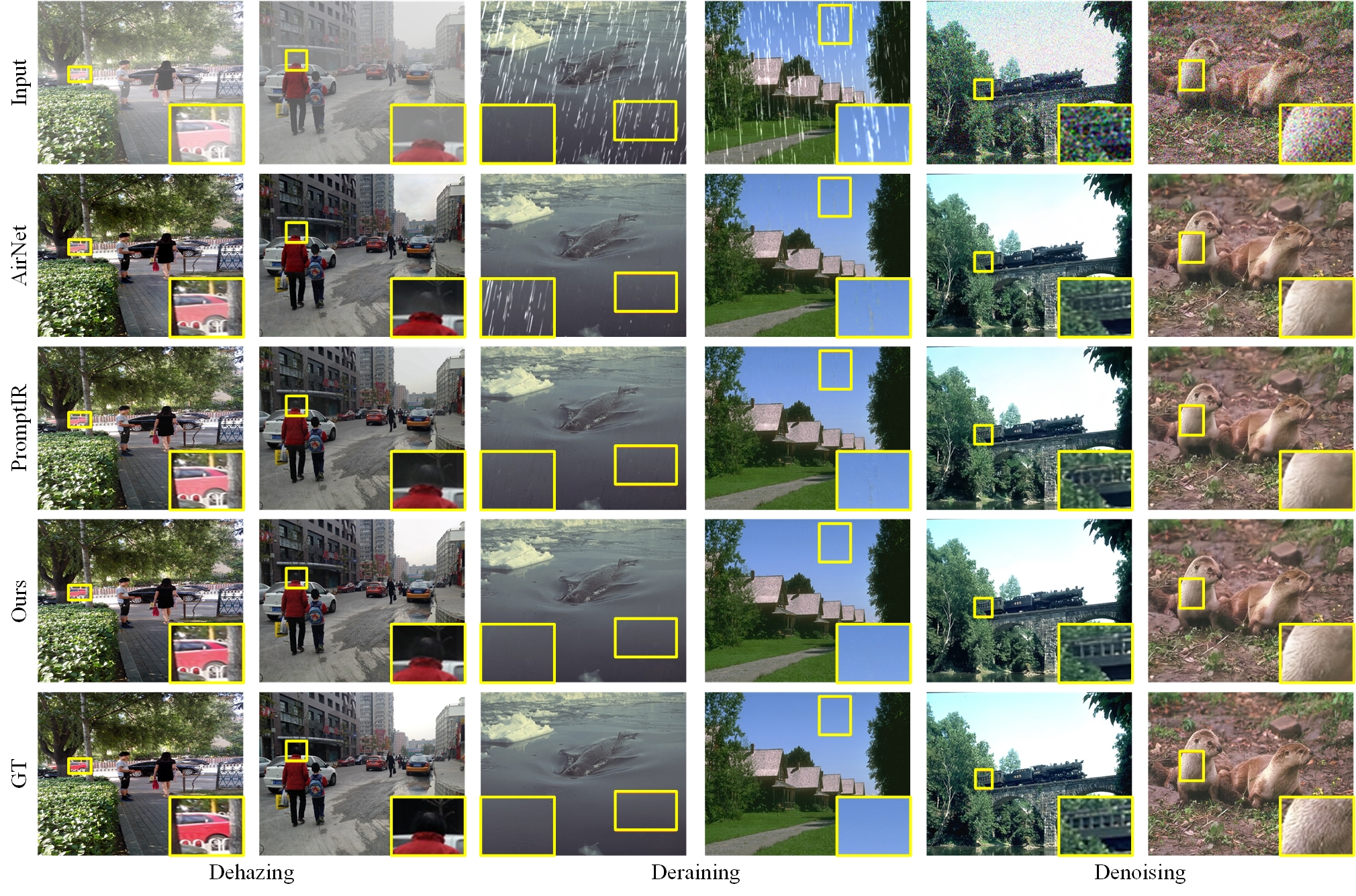}
	 \vspace{-0.3cm}
	\caption{Comparison of visual effects in restored results using all-in-one methods trained on multi-task settings. }\vspace{-0.3cm}
	\label{fig6}
\end{figure*}

\begin{table*}[t!]
	\centering
	\caption{Performance Comparison of Different Methods on Three Degradation Tasks. The best values are highlighted in bold.}
	\renewcommand\arraystretch{1.2}
	\fontsize{10}{10}\selectfont  
	\begin{small}
		\begin{tabular}{c|cc|cc|cc|cc|cc|cc}
			\hline
			\multirow{3}{*}{Methods} & \multicolumn{2}{c|}{Dehazing} & \multicolumn{2}{c|}{Deraining} & \multicolumn{6}{c|}{Denoising} & \multicolumn{2}{c}{Average} \\ \cline{2-13}
			& \multicolumn{2}{c|}{SOTS} & \multicolumn{2}{c|}{Rain100L} & \multicolumn{2}{c|}{BSD68$_{\sigma=15}$} & \multicolumn{2}{c|}{BSD68$_{\sigma=25}$} & \multicolumn{2}{c|}{BSD68$_{\sigma=50}$} & \multicolumn{2}{c}{All Tasks} \\ \cline{2-13}
			& PSNR & SSIM & PSNR & SSIM & PSNR & SSIM & PSNR & SSIM & PSNR & SSIM & PSNR & SSIM \\ \hline
			BRDNet \cite{48} & 23.23 & 0.895 & 27.42 & 0.895 & 32.26 & 0.898 & 29.76 & 0.836 & 26.34 & 0.693 & 27.80 & 0.843 \\ 
			LPNet \cite{49} & 20.84 & 0.828 & 24.88 & 0.784 & 26.47 & 0.778 & 24.77 & 0.748 & 21.26 & 0.552 & 23.64 & 0.738 \\ 
			FDGAN \cite{61} & 24.71 & 0.929 & 29.89 & 0.933 & 30.25 & 0.910 & 28.81 & 0.868 & 26.43 & 0.776 & 28.02 & 0.883 \\ 
			DL \cite{52} & 26.92 & 0.931 & 32.62 & 0.931 & 33.05 & 0.914 & 30.41 & 0.861 & 26.90 & 0.740 & 29.98 & 0.876 \\ 
			MPRNet \cite{62} & 25.28 & 0.955 & 33.57 & 0.954 & 33.54 & 0.927 & 30.89 & 0.880 & 27.56 & 0.779 & 30.17 & 0.899 \\ 
			AirNet \cite{50} & 27.94 & 0.962 & 34.90 & 0.967 & 33.92 & 0.933 & 31.26 & 0.888 & 28.00 & 0.797 & 31.20 & 0.910 \\ 
			PromptIR \cite{37} & 30.58 & 0.974 & 36.37 & 0.972 & 33.98 & 0.933 & 31.31 & 0.888 & 28.06 & 0.799 & 32.06 & 0.913 \\ 
			Ours & \textbf{31.61} & \textbf{0.981} & \textbf{39.00} & \textbf{0.985} & \textbf{34.12} & \textbf{0.935} & \textbf{31.46} & \textbf{0.892} & \textbf{28.19} & \textbf{0.803} & \textbf{32.85} & \textbf{0.919} \\ \hline
		\end{tabular}\label{lab1}\vspace{-0.3cm}
	\end{small}
\end{table*}
\section{Experiments}\label{R4}
\subsection{Datasets}
Following the approach of previous research, we used corresponding datasets for different restoration tasks to validate the performance of the model. Specifically: For image dehazing, we selected the SOTS subset from the RESIDE (outdoor) dataset \cite{53}, which comprised 72,135 training images and 500 test images;  For image deraining, we used the Rain100L dataset \cite{54}, which included 200 clean-rain image pairs for model training and an additional 100 pairs for testing; For image denoising, we jointly used the BSD400 \cite{55} and WED \cite{56} datasets for model training. The training set contained 5,144 clear images, from which we generated noisy images by adding Gaussian noise with standard deviations of 5, 25, and 50. The trained model was tested on the BSD68 dataset \cite{57}, which contains 68 clear images with noisy images generated by adding Gaussian noise with standard deviations of 15, 25, and 50; For deblurring, we used the GoPro dataset \cite{58} for motion deblurring, which included 2,103 training images and 1,111 test images; For low-light image enhancement, we used the LOL-v1 dataset \cite{59}, which contained 485 training images and 15 test images.

\subsection{Implementation Details and Evaluation Metrics}
All experiments in this paper were conducted on a single NVIDIA GeForce RTX 4090 GPU, and the model was implemented using the PyTorch 1.12.0 framework. During the training phase, we used the Adam optimizer to optimize the network, setting the initial learning rate to $2 \times 10^{-4}$ and adjusting it using the cosine annealing strategy. Additionally, we randomly cropped the images to a size of $128 \times 128$ pixels for training. In each small batch, data augmentation was performed by flipping the images horizontally or vertically to expand the training sample size. Under the All-in-One setting, we merged these datasets and trained a single model under three and five degradation settings, respectively. The training process lasted for 150 epochs, and the model was directly tested across multiple restoration tasks. Under the single-task setting, we trained individual models for each specific restoration task, with each model trained for 150 epochs and tested on its respective test set. In experiments, the number of experts $K$ and the hyperparameter $\ell_{balance}$ were heuristically set to 2 and 0.0001, respectively. To objectively evaluate the quality of the restoration results, we adopted commonly used image quality assessment metrics, namely Peak Signal-to-Noise Ratio (PSNR) and Structural Similarity Index Measure (SSIM), to assess the quality of the reconstructed results.
\begin{table*}[t]
	\centering
	\caption{Performance Comparison of Different Methods on Five Image Restoration Tasks. The best values are highlighted in bold. Denoising results are reported at a noise level with a standard deviation of 25.}
	\renewcommand\arraystretch{1.2}
	\fontsize{10}{10}\selectfont  
	\begin{small} 
		\begin{tabular}{c|cc|cc|cc|cc|cc|cc}
			\toprule
			\multirow{2}{*}{Methods} & \multicolumn{2}{c|}{Dehazing} & \multicolumn{2}{c|}{Deraining} & \multicolumn{2}{c|}{Denoising} & \multicolumn{2}{c|}{Deblurring} & \multicolumn{2}{c|}{Low-Light} & \multicolumn{2}{c}{Average} \\ \cline{2-13}
			& PSNR & SSIM & PSNR & SSIM & PSNR & SSIM & PSNR & SSIM & PSNR & SSIM & PSNR & SSIM \\
			\midrule
			NAFNet \cite{63} & 25.23 & 0.939 & 35.56 & 0.967 & 31.02 & 0.883 & 26.53 & 0.808 & 20.49 & 0.809 & 27.76 & 0.881 \\
			HINet \cite{64}& 24.74 & 0.937 & 35.67 & 0.969 & 31.00 & 0.881 & 26.12 & 0.788 & 19.47 & 0.800 & 27.40 & 0.875 \\
			DGUNet \cite{65}& 24.78 & 0.940 & 36.62 & 0.971 & 31.10 & 0.883 & 27.25 & 0.837 & 21.87 & 0.823 & 28.32 & 0.891 \\
			MIRNetV2 \cite{66}& 24.03 & 0.927 & 33.89 & 0.954 & 30.97 & 0.881 & 26.30 & 0.799 & 21.52 & 0.815 & 27.34 & 0.875 \\
			SwinIR \cite{22}& 21.50 & 0.891 & 30.78 & 0.923 & 30.59 & 0.868 & 24.52 & 0.773 & 17.81 & 0.723 & 25.04 & 0.835 \\
			Restormer \cite{23}& 24.09 & 0.927 & 34.81 & 0.962 & 31.49 & 0.884 & 27.22 & 0.829 & 20.41 & 0.806 & 27.60 & 0.881 \\
			DL \cite{52}& 20.54 & 0.826 & 21.96 & 0.762 & 23.09 & 0.745 & 19.86 & 0.672 & 19.83 & 0.712 & 21.05 & 0.743 \\
			Transweather \cite{67}& 21.32 & 0.885 & 29.43 & 0.905 & 29.00 & 0.841 & 25.12 & 0.757 & 21.21 & 0.792 & 25.22 & 0.836 \\
			TAPE \cite{68} & 22.16 & 0.861 & 29.67 & 0.904 & 30.18 & 0.855 & 24.47 & 0.763 & 18.97 & 0.621 & 25.09 & 0.801 \\
			AirNet \cite{50}& 21.04 & 0.884 & 32.98 & 0.956 & 31.20 & 0.897 & 24.35 & 0.781 & 18.18 & 0.735 & 25.49 & 0.846 \\
			IDR \cite{30}& 25.24 & 0.943 & 35.63 & 0.965 & 31.60 & 0.887 & 27.87 & 0.846 & 21.34 & 0.826 & 28.34 & 0.893 \\
			Ours & \textbf{31.05} & \textbf{0.980} & \textbf{38.32} & \textbf{0.982} & \textbf{31.40} & \textbf{0.888} & \textbf{29.41} & \textbf{0.890} & \textbf{23.00} & \textbf{0.845} & \textbf{30.64} & \textbf{0.917} \\
			\bottomrule
		\end{tabular}\label{lab2}
	\end{small}
\end{table*}
  \begin{figure*}[!t]
  	\centering
  	\includegraphics[width=0.92\linewidth]{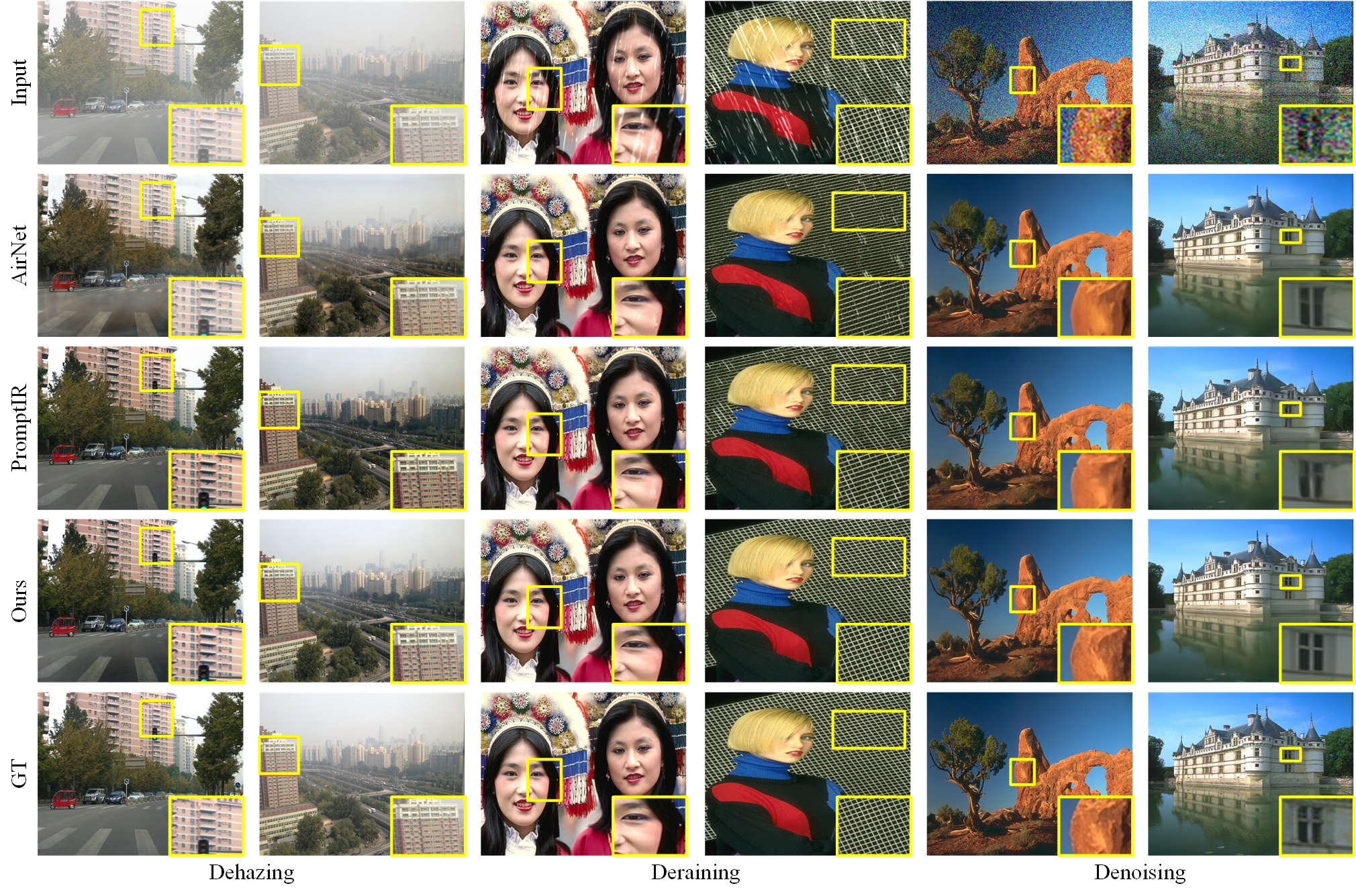}
  	\vspace{-0.3cm}
  	\caption{Comparison of visual effects in restored results using all-in-one methods trained on single-task restoration settings. }\vspace{-0.3cm}
  	\label{fig7}
  \end{figure*}

\subsection{Comparison with State-of-the-art Methods}
\subsubsection{Comparison of Restoration Across Three Degradations}
We conducted a comprehensive performance evaluation of our method on three typical image restoration tasks: image dehazing, deraining, and denoising. To further assess its restoration effectiveness, we compared it with various advanced image restoration methods, including general methods such as BRDNet \cite{48}, LPNet \cite{49}, FDGAN \cite{61}, and MPRNet \cite{62}, as well as specialized All-in-One image restoration methods like DL \cite{52}, AirNet \cite{50}, and PromptIR \cite{37}. To demonstrate the differences in visual effects of the restored images, we have shown the visual results of some images after dehazing, deraining, and denoising in Fig.\ref{fig6}. It is worth noting that the general image restoration methods, including BRDNet, LPNet, FDGAN, and MPRNet, do not provide trained parameters and therefore cannot display the visual effects of their restoration results. Hence, following the PromptIR processing mode, we only compared our proposed method with these methods based on objective evaluation results. From the results shown in Fig.\ref{fig6}, it can be seen that the proposed method can more effectively restore the color information of the source image during the dehazing process. During the rain removal process, it not only effectively removes rain streaks but also restores lost image information. In the denoising process, it can more effectively preserve the edge texture details in the source image, making the restoration result closer to the ground truth (GT).
\begin{table*}[h]
	\centering
	\caption{Comparison of Dehazing Performance of Different Methods on the SOTS-Outdoor Dataset in Single-Task Settings}
		\renewcommand\arraystretch{1.2}
	\fontsize{10}{10}\selectfont 
	\begin{small} 
	\begin{tabular}{c|c|c|c|c|c|c|c|c}
		\hline
		Metrics & MSCNN \cite{70}& AODNet \cite{71} & EPDN \cite{72}& FDGAN \cite{61} & Restormer \cite{23} & AirNet \cite{50} & PromptIR \cite{37}& Ours \\
		\hline
		PSNR & 22.06 & 20.29 & 22.57 & 23.15 & 30.87 & 23.18 & 31.31 & \textbf{31.83} \\
		SSIM & 0.908 & 0.877 & 0.863 & 0.921 & 0.969 & 0.900 & 0.973 & \textbf{0.981} \\
		\hline
	\end{tabular}
\end{small}
\label{lab3}\vspace{-0.3cm}
\end{table*}
\begin{table*}[ht]  
	\centering  
	\caption{Comparison of Deraining Performance of Different Methods on the Rain100L Dataset in Single-Task Settings}  
	\renewcommand\arraystretch{1.2}
	\fontsize{10}{10}\selectfont  
	\begin{small}
		\begin{tabular}{l|c|c|c|c|c|c|c|c}  
			\hline  
			Metrics & UMR \cite{73}& SIRR \cite{74}& MSPFN \cite{75}& LPNet\cite{60} & Restormer \cite{23} & AirNet \cite{50}& PromptIR\cite{37} & Ours \\  
			\hline
			PSNR    & 32.39 & 32.37 & 33.50 & 33.61 & 36.74    & 34.90  & 37.04    & \textbf{39.06}  \\  
			SSIM    & 0.921 & 0.926 & 0.948 & 0.958 & 0.978    & 0.977  & 0.979    & \textbf{0.985}  \\  
			\hline 
		\end{tabular}\label{lab4}\vspace{-0.3cm}	
	\end{small}
\end{table*}  

To objectively evaluate the restoration performance of different methods, we used PSNR and SSIM to measure the results generated by our method. The results of other comparative methods were obtained from the data provided in the PromptIR experiment. As shown in Table \ref{lab1}, our method exhibits superior performance in improving restoration quality compared to other methods. Specifically, our method improved the average PSNR by 0.79 dB compared to the high-performance PromptIR method and by 1.65 dB compared to the suboptimal AirNet. In detail, compared with PromptIR, our method achieved a performance improvement of 2.63 dB in the deraining task and 1.03 dB in the dehazing task. In the denoising task, our method improved performance by 0.14 dB, 0.15 dB, and 0.13 dB for noise levels of 15, 25, and 50, respectively. In terms of SSIM, our method also shows advantages compared to PromptIR, with improvements of 0.013 in the deraining task and 0.007 in the dehazing task. For denoising tasks with noise levels of 15, 25, and 50, our method improved SSIM by 0.002, 0.004, and 0.004, respectively. These results demonstrate the effectiveness and superiority of the method proposed in this paper.
\subsubsection{Comparison of Restoration Across Five Degradations}
To comprehensively verify the performance of the method proposed in this paper, we followed the design principles of existing methods and applied our method to five different image restoration tasks: image dehazing, image deraining, image denoising, image deblurring, and low-light image restoration. However, since existing methods do not provide testable parameters under this setting, we are unable to directly display a visual comparison of the recovery results of different methods. Similar to the PromptIR method, our comparison is primarily based on objective evaluation results. As shown in Table \ref{lab2}, our method demonstrates significant advantages in handling these various image restoration tasks, further proving the rationality, progressiveness, and applicability of our method.\vspace{-0.2cm}
\begin{figure*}[t!]
	\centering
	\includegraphics[width=0.8\linewidth]{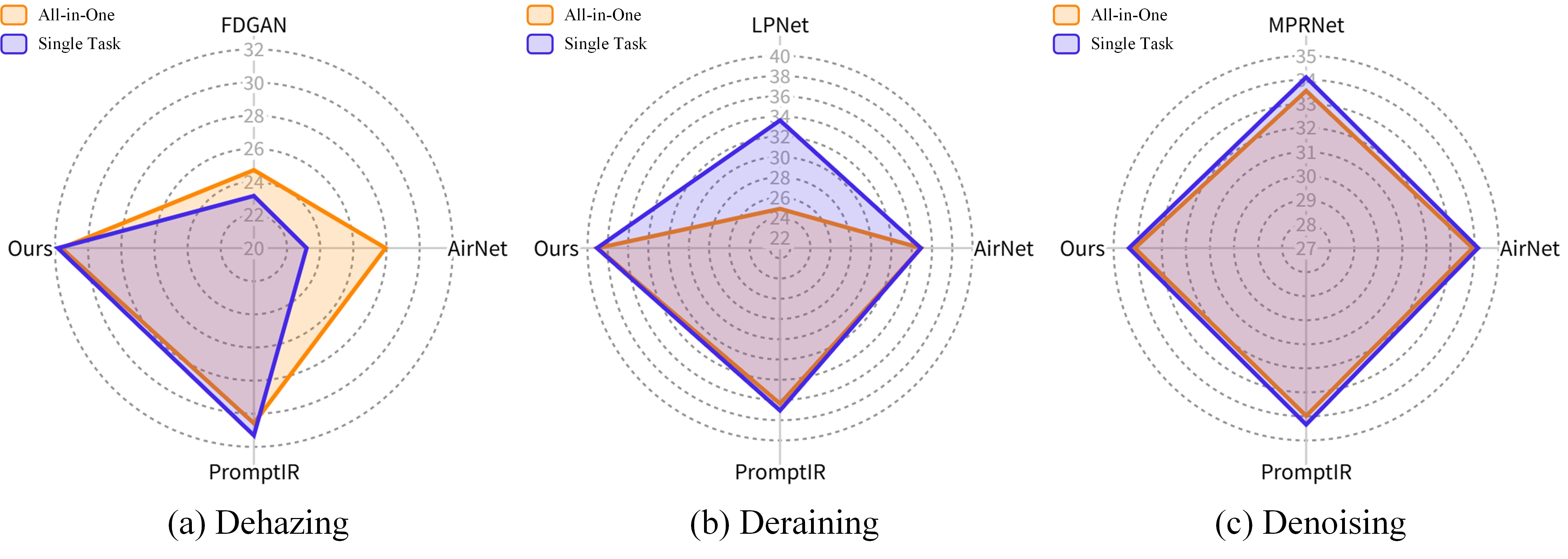}
	\vspace{-0.3cm}
	\caption{Performance changes of different All-in-One methods under All-in-One and single-task restoration settings.}\vspace{-0.3cm}
	\label{fig8}
\end{figure*}
\subsection{Comparison on Single-Task Restoration Settings}
The design of this method comprehensively considers the specific requirements of single tasks on the network. This ensures that the performance of single-task image restoration is not sacrificed within the All-in-One framework to achieve coordination of multiple tasks. To verify this, we trained our model on three challenging datasets: SOTS-Outdoor, Rain100L, and BSD68, respectively. We then comprehensively compared its performance with several other representative methods on different testing tasks. The restoration results shown in Fig.\ref{fig7} clearly indicate that our method exhibits stronger restoration ability compared to other methods, achieving the best visual effect. Meanwhile, these restoration results did not introduce any significant artifacts or false information, further demonstrating the effectiveness of our method. Additionally, the objective evaluation data in Tables \ref{lab3}-\ref{lab5} fully confirm that the performance of our method is still significantly better than that of the relevant comparative methods in single-task mode.

To comprehensively evaluate the performance of our method in the All-in-One setting and compare it with the single-task setting, we present detailed performance changes of different All-in-One methods under both settings in Fig.\ref{fig8}. Through comparative analysis, it is evident that our method maintains a high degree of stability, showing no significant fluctuations whether in the All-in-One or single-task recovery setting. This excellent performance is primarily due to our method's ability to effectively coordinate the requirements of different tasks for the network framework. Additionally, it significantly mitigates potential performance trade-offs when handling multiple tasks. This demonstrates that the proposed method has stronger generalization ability and practical application value compared to the comparison methods, maintaining excellent performance in various settings.

\begin{table*}[ht]
	\centering
	\caption{Comparison of Denoising Performance of Different Methods in Single-Task Settings on the BSD68 Dataset}
	   \renewcommand\arraystretch{1.2}
	\fontsize{10}{10}\selectfont  
	\begin{small}
	\begin{tabular}{l|c|c|c|c|c|c|c|c}
		\hline
		Settings & Metrics & DnCNN \cite{8}& IRCNN \cite{76}& FFDNet \cite{9}& BRDNet \cite{48}& AirNet \cite{50}& \textbf{Ours} \\
		\hline
	\multirow{2}{*}{BSD68$_{\sigma=15}$}	& PSNR & 33.89 & 33.87 & 33.87 & 34.10 & 34.14 & \textbf{34.36} \\
		 & SSIM & 0.930 & 0.929 & 0.929 & 0.929 & 0.936 & \textbf{0.938} \\
		\hline
		\multirow{2}{*}{BSD68$_{\sigma=25}$} & PSNR & 31.23 & 31.18 & 31.21 & 31.43 & 31.48 & \textbf{31.73} \\
		& SSIM & 0.883 & 0.882 & 0.882 & 0.885 & 0.893 & \textbf{0.898} \\
		\hline
		\multirow{2}{*}{BSD68$_{\sigma=50}$} & PSNR & 27.92 & 27.88 & 27.96 & 28.16 & 28.23 & \textbf{28.50} \\
		& SSIM & 0.789 & 0.790 & 0.789 & 0.794 & 0.806 & \textbf{0.814} \\
		\hline
	\end{tabular}\label{lab5}\vspace{-0.3cm}
	\end{small}
\end{table*}

\subsection{Ablation Study}
The proposed method mainly consists of four core components: TSPG, MESE, FD, and MEE. To evaluate the effectiveness of each component, we performed ablation studies on each module of the method. All experiments were conducted under the All-in-One setting, covering three degradation scenarios: image dehazing, image deraining, and image denoising. The quantitative evaluation results (mean values) of these ablation experiments are presented in Table \ref{lab6}.

\textbf{Effectiveness of TSPG}:
TSPG is primarily used to generate task-specific prompts, which help in more accurately selecting experts suitable for the current task. To verify the effectiveness of TSPG, we conducted experiments without the MESE and FD-MEE components. Specifically, we tested a method where the output features from TSPG were directly concatenated with the output features from the convolutional layer in MESE and then processed through the Transformer layer. The data in Table \ref{lab6} clearly show that adding TSPG significantly improved model performance. This enhancement is mainly due to TSPG can provide detailed descriptions of image degradation at a global level, thereby increasing the representation of features and substantially improving image restoration quality.
\begin{table}[ht]
	\centering
	\caption{Ablation studies for each component. These data represent the average PSNR and SSIM across all images in the three tasks.}
		   \renewcommand\arraystretch{1.3}
	\fontsize{10}{10}\selectfont  
	\begin{small}
	\begin{tabular}{c|c|c|c|c|c}
		\hline
		\multicolumn{2}{c|}{STP-G-MESE} & \multicolumn{2}{c|}{FD-MEE} & \multicolumn{2}{c}{Average} \\
		\hline  
		TSPG & MESE & FD & MEE & PSNR & SSIM \\
		\hline
		$\times$ & $\times$ & $\times$ & $\times$ & 29.97 & 0.865 \\
		$\checkmark$ & $\times$ & $\times$ & $\times$ & 31.15 & 0.883 \\
		$\checkmark$ & $\checkmark$ & $\times$ & $\times$ & 32.21 & 0.904 \\
		$\checkmark$ & $\checkmark$ & $\checkmark$ & $\times$ & 32.64 & 0.910 \\
		$\checkmark$ & $\checkmark$ & $\times$ & $\checkmark$ & 32.57 & 0.907 \\
		$\checkmark$ & $\checkmark$ & $\checkmark$ & $\checkmark$ & 32.85 & 0.919 \\
		\hline
	\end{tabular}\label{lab6}\vspace{-0.3cm}
\end{small}
\end{table}

\textbf{Effectiveness of MESE}: 
MESE is mainly used to select and ensemble experts based on the results output by TSPG. According to the data in Table \ref{lab6}, we can observe a significant improvement in model performance when both MESE and TSPG are utilized. This observation demonstrates the positive role of MESE and TSPG in the expert selection process. Furthermore, it confirms the crucial role of the expert selection and ensemble mechanisms within MESE in enhancing overall performance.

\textbf{Effectiveness of FD}:
FD is primarily utilized to separate features, enabling the independent processing of high-frequency and low-frequency components. The data in Table \ref{lab6} shows that the model's performance improves when these components are processed separately. This result highlights the importance of decomposing the features into their low-frequency and high-frequency components.

\textbf{Effectiveness of MEE}:
MEE is used to integrate the outputs of multiple experts at a global level. The results in Table \ref{lab6} show that incorporating MEE leads to further improvement in model performance. This indicates that MEE has a positive impact within the overall network framework.
\begin{figure}[!t]
	\centering
	\includegraphics[width=0.9\linewidth]{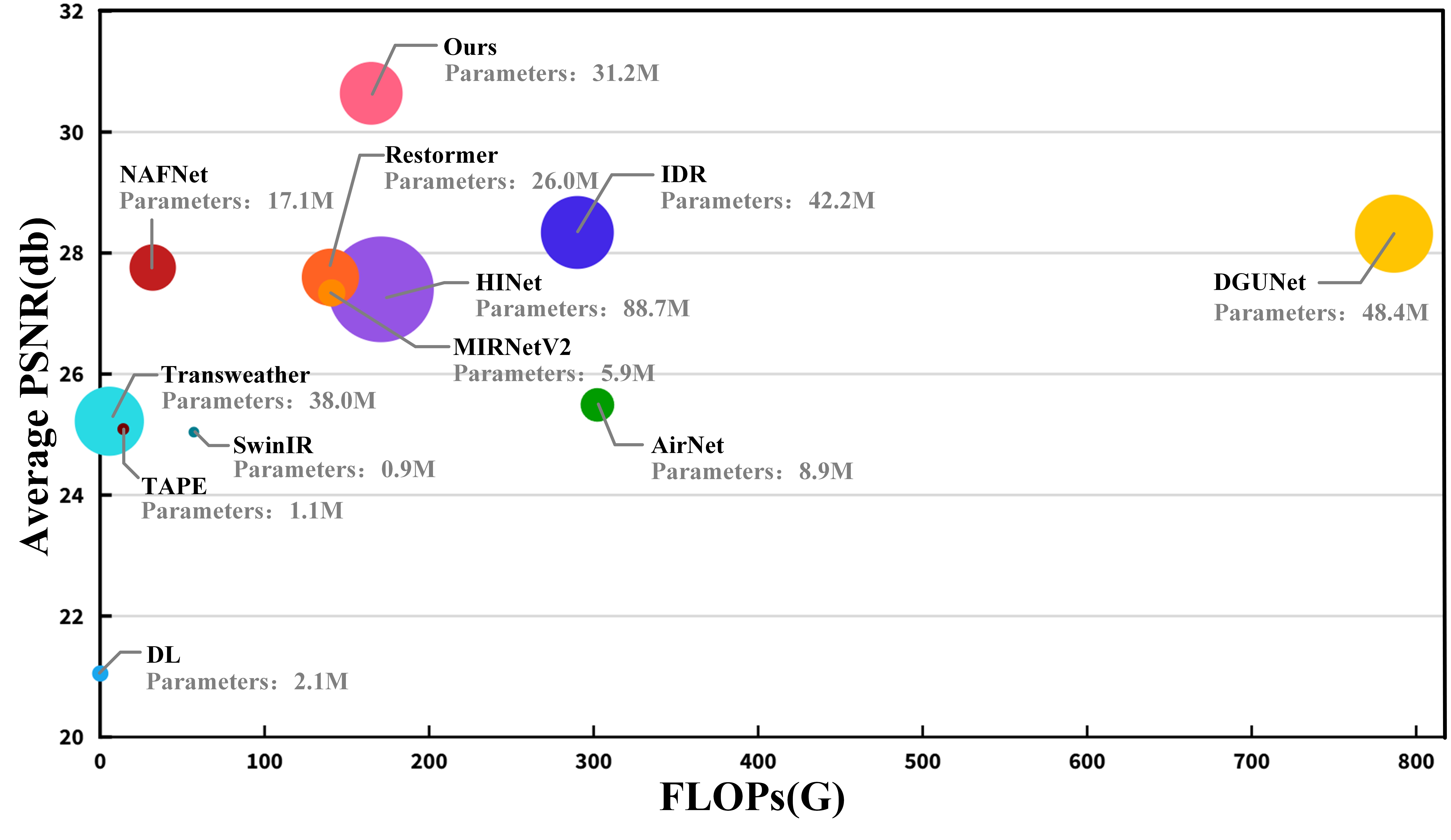}
	\vspace{-0.3cm}
	\caption{Comparison of parameter scales and model performance: This graph plots the relationship between FLOPs (measured on 256$\times$256$\times$3 images) and PSNR  evaluation results. The size of each bubble (circle area) represents the number of parameters in each model.}\vspace{-0.3cm}
	\label{fig9}
\end{figure}

\subsection{Limitation and Future Work}
As depicted in Fig. \ref{fig9}, the proposed method demonstrates strong restoration performance and low model complexity. However, its parameter size remains relatively large compared to certain CNN-based methods, such as TAPE, DL, and AirNet, due to the inclusion of the Transformer architecture. Nevertheless, our method’s parameter size is notably lower compared to other Transformer-based methods like IDR, Restormer, and DGUNet. In future work, we plan to explore strategies to enhance the restoration performance while maintaining the model’s lightweight characteristics.

\section{Conclusion} \label{R5}
In this paper, we propose a multi-expert adaptive selection mechanism designed to address the diverse requirements of various tasks within the field of image restoration. We develop a feature representation method that captures image information at both the single-pixel channel level and the global level, including low-frequency and high-frequency components. This method establishes a solid foundation for expert selection and ensemble. Our multi-expert selection and ensemble scheme adapts to the content characteristics of the input image and specific task prompts, ensuring the selection of the most suitable experts from our expert library. This approach not only meets the unique demands of different tasks but also maintains an optimal balance, allowing tasks to share expert resources without interference. Furthermore, by selecting experts based on image content, our mechanism promotes knowledge sharing and learning across tasks, ultimately enhancing overall restoration performance. Experimental results confirm that our proposed method outperforms existing approaches, demonstrating superior restoration performance and generalization ability in multi-task image restoration scenarios.

\bibliographystyle{IEEEtran}
\bibliography{mybibfile1}
\begin{IEEEbiography}[{\includegraphics[width=1.2in,height=1.30in,clip,keepaspectratio]{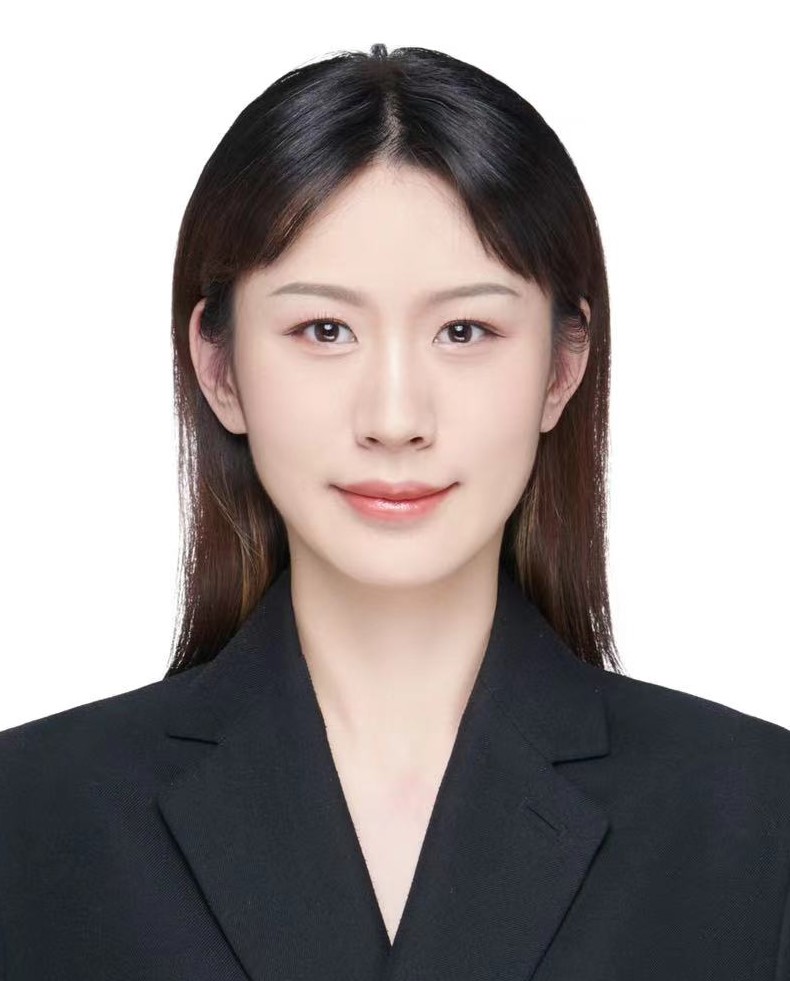}}]{Xiaoyan Yu}  is currently a Ph.D. candidate in the School of Computer Science and Technology, Beijing Institute of Technology. Her research interests include social event mining, natural language processing and image processing.	
\end{IEEEbiography}
\begin{IEEEbiography}[{\includegraphics[width=1.2in,height=1.30in,clip,keepaspectratio]{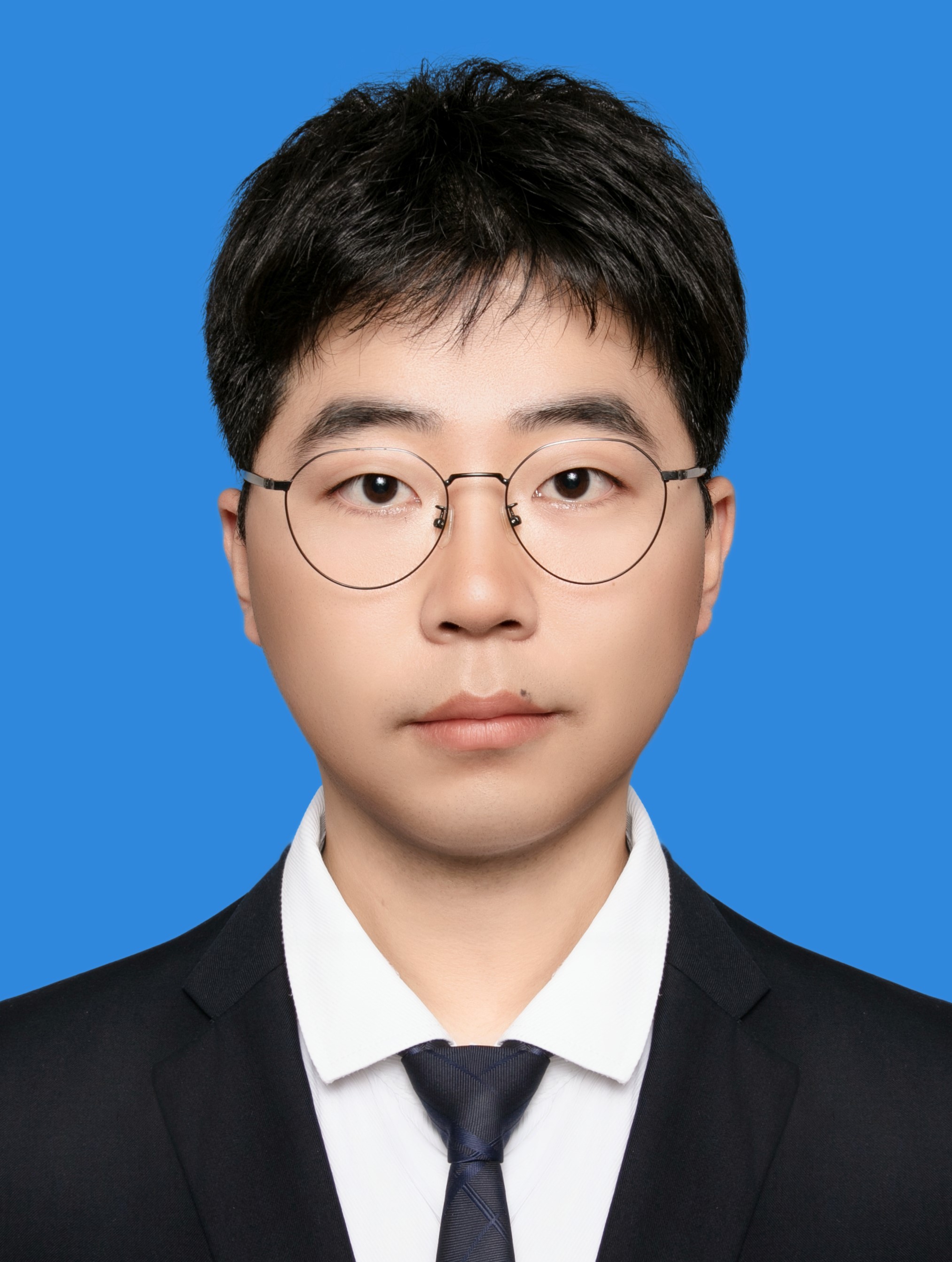}}]{Shen Zhou}  received his B.E. degree in Communication Engineering from Suqian University in 2021. He is currently pursuing his M.E. degree in Communication Engineering at the School of Information Engineering and Automation, Kunming University of Science and Technology. His research interests include image processing and computer vision.
\end{IEEEbiography}
\begin{IEEEbiography}[{\includegraphics[width=1.2in,height=1.30in,clip,keepaspectratio]{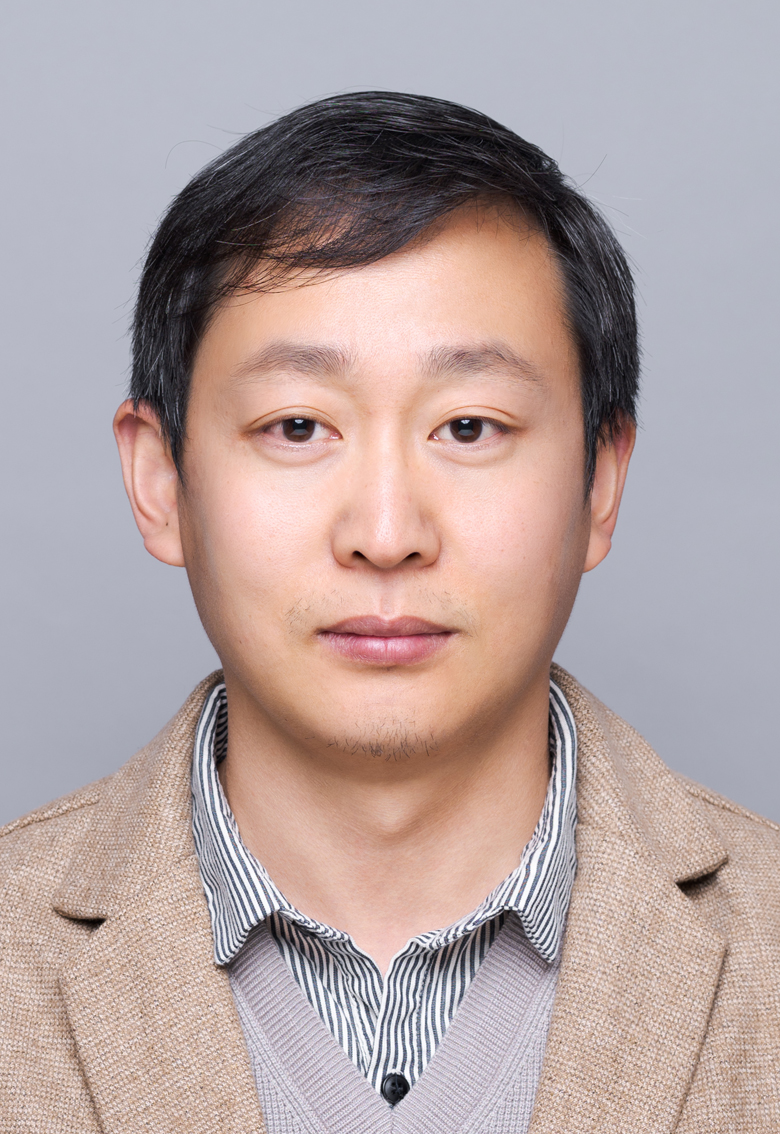}}]{Huafeng Li}  received the M.S. degrees in applied mathematics major from Chongqing University in 2009 and obtained his Ph.D. degree in control theory and control engineering major from Chongqing University in 2012.  He is currently a professor at the School of Information Engineering and Automation, Kunming University of Science and Technology, China. His research interests include image processing, computer vision, and information fusion. He has authored or coauthored more than 50 scientific articles in CVPR, IJCV, AAAI, ACMMM, ICME, IEEE TIP, IEEE TIFS, IEEE TNNLS, IEEE TMM, IEEE TCSVT, IEEE TGRS, IEEE TCI, IEEE TII, IEEE TETCI, IEEE TITS, IEEE TIM, IEEE/CAA JAS, PR, INFFUS, NeuNet, INS, ESWA, KBS, etc. 
\end{IEEEbiography}
\begin{IEEEbiography}[{\includegraphics[width=1.2in,height=1.30in,clip,keepaspectratio]{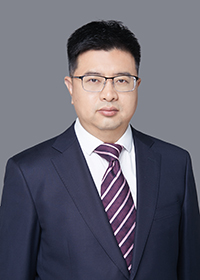}}]{Liehuang Zhu} (Senior Member, IEEE) received the Ph.D. degree in computer science from the Beijing Institute of Technology, Beijing, China, in 2004. He is currently a Professor with the School of Cyberspace Science and Technology, Beijing Institute of Technology. He is selected into the Program for New Century Excellent Talents in University from the Ministry of Education, China. He has published over 100 SCI-indexed research papers in these areas, including ten more IEEE/ACM TRANSACTIONS PAPERS. His research interests include blockchain, Internet of Things, data security, and artificial intelligence security. Prof. Zhu has been granted a number of IEEE Best Paper Awards, including IWQoS’17 and TrustCom’18.
\end{IEEEbiography}
\end{document}